\documentclass[11pt,oneside,a4paper]{article}
\usepackage{graphicx}
\usepackage{setspace}
\usepackage{lineno}
\usepackage{natbib}
\usepackage[justification=centering]{caption}
\usepackage{color,soul} 
\title{Fruit Detection, Segmentation and 3D Visualisation of Environments in Apple Orchards}
\date{}
\author{Hanwen Kang and Chao Chen$^{*}$}

\begin{document} 
	\maketitle
	\setcitestyle{notesep={; },round,aysep={,},yysep={;}}
	
	\begin{abstract}
		Robotic harvesting of fruits in orchards is a challenging task, since high density and overlapping of fruits and branches can heavily impact the success rate of robotic harvesting. Therefore, the vision system is demanded to provide comprehensive information of the working environment to guide the manipulator and gripping system to successful detach the target fruits. In this study, a deep learning based one-stage detector DaSNet-V2 is developed to perform the multi-task vision sensing in the working environment of apple orchards. DaSNet-V2 combines the detection and instance segmentation of fruits and semantic segmentation of branch into a single network architecture. Meanwhile, a light-weight backbone network LW-net is utilised in the DaSNet-V2 model to improve the computational efficiency of the model. In the experiment, DaSNet-V2 is tested and evaluated on the RGB-D images of the orchard. From the experiment results, DaSNet-V2 with light-weight backbone achieves 0.844, 0.858, and 0.795 on the $F_1$ score of the detection, and mean intersection of union on the instance segmentation of fruits and semantic segmentation of branches, respectively. To provide a direct-viewing of the working environment in orchards, the obtained sensing results are illustrated by 3D visualisation
		. The robustness and efficiency of the DaSNet-V2 in detection and segmentation are validated by the experiments in the real-environment of apple orchard.
	\end{abstract}

	\section{Introduction}
	Nowadays, with the increasing cost and difficulty in availability of the labour resource \citep{Aus_report}, agriculture industry requires transformation from the labour-intensive industry to the technology-intensive industry. Robotic technology has shown a promising prospect in terms of improving the efficiency and yield of agriculture production. Different from the traditional autonomous equipment which has been widely applied in the harvesting of commercial crops such as wheat and soybean, to design a robotic system for automatic harvesting of fruits is a more challenge task \citep{2019harvesting_robot}. The vision system is the key to the harvesting since it senses the working environment and guides the manipulator to detach fruits. Moreover, due to the complex conditions in fruit orchards, such as 'free growth' and densely arranged branches and fruits, fruit harvesting robots are required to understand the working environment to increase the rate of success during the operation \citep{2016TMW_review}. Meanwhile, other environmental factors, including illumination variance and occlusion, can also heavily affect the performance of the vision system.   
	
	This work developed a multi-function Deep Convolution Neural Network DaSNet-V2 to perform the vision sensing of working environment in apple orchards. DaSNet-V2 adopts one-stage detector architecture to perform the detection and instance segmentation of fruits. Meanwhile, a semantic segmentation branch is grafted to the network to segment branches in orchards. To ensure the computational availability of the network model on the embedded computing devices, a light-weight backbone based on the residual network is developed and utilised in the DaSNet-V2. DaSNet-v2 is evaluated in the data which is collected from apple orchards, 3D visualisation of processed orchard environments by means of the DaSNet-v2 is also illustrated in the experiments.
	
	The rest of paper is organised as follow. Section 2 reviews the related works. Sections 3 and 4 introduce the methodology and experiment of the work, respectively. In section 5, the conclusions and future work are presented.
	
	\section{Literature Review}
	Vision sensing in fruit orchards has been extensively studied in previous studies. There are currently two classes of approaches: traditional machine learning based algorithm and deep learning based algorithms. Traditional machine learning based algorithms use image features extracted from the 2D image space or 3D geometry space and machine learning based classifier to classify, detect, and segment the elements in the images \citep{2012traditional_machine_learning}. A number of work of applying traditional machine learning based algorithm on vision sensing in agriculture environment. \cite{2016TMW_1} applied descriptor of colour and geometry features to describe the appearance of red apples, and a clustering algorithm based on Euclidean distance in feature space to segment and detect the fruits in the images. The similar processing techniques of performing segmentation and detection in vision sensing in orchard environment are also presented in the works of \citep{2012TMW_2,2016TMW_3,2019TMW_4,2018TMW_5}. \cite{2018TMW_6} applied multiple image features and Latent Dirichlet Allocation (LDA) model to perform unsupervised segmentation of the plant and fruits in the greenhouse environment, showing that traditional machine learning based algorithm can be used in generating labelling data for deep learning based algorithms. More approaches which applied traditional machine learning based algorithm to perform vision sensing in orchards can also be found in the reviews of \citep{2012TMW_review2} and \citep{2016TMW_review}.
	
	Deep learning based algorithm is developed more recently. Compared to the traditional machine learning based algorithm, deep learning based algorithm achieved superior accuracy and generalisation ability in classification, detection, and segmentation \citep{2018deep_survey}. Deep learning based detection algorithm can be classified into two class: two-stage detector and one-stage detector \citep{2017focal}. The representative work of the two-stage detector is Region Convolution Neural Network (RCNN), which include Fast/Faster-RCNN \citep{2015fast-rcnn,2015faster-rcnn} and Mask-RCNN \citep{2017mask-rcnn}. Faster-RCNN applies Region Proposal Network (RPN) and Region of Interest (RoI) pooling to combine the RoI searching and classification into a single network architecture, which increases the computational efficiency of the model. Mask-RCNN further combines mask segmentation branch into the model, which allows the network to segment the corresponding area for each object within the images. The representative work of the one-stage detector is YOLO \citep{2018yolov3}. Different from RCNN predict the possible RoI from feature maps, YOLO predicts the object on each grid of feature maps. Compared to the RCNN, YOLO achieves equally performance with much improved computational efficiency. Also, Single Pixel Reconstruction Network (SPRNet) \citep{2019SPRNet}, which combines the instance segmentation branch into the architecture of the one-stage detector, achieving the same function compared to the Mask-RCNN. The work of \citep{2016DLW_1} and \citep{2017DLW_2} applied Faster-RCNN in the detection of fruits, accurate detection performance was reported from both works. \cite{2019DLW_3} applied Mask-RCNN for strawberry harvesting robot in a non-structured environment. \cite{2019DLW_4} and \cite{2019DLW_5} applied YOLO-V3 in the monitoring of fruit growth in the apple orchard and mango orchard. \cite{2019DasNet-v1} combined the semantic segmentation and detection into a one-stage detector, to perform the fruit detection and branch segmentation in the apple orchard for robotic harvesting. Other deep learning based algorithms such as Fully Convolution Network (FCN) \citep{2015FCN} are also being studied and applied in performing vision sensing in the agriculture environments, such as the works of \citep{2019DLW_6} and \citep{2019DLW_7}. More works of using deep learning based algorithm in the in-field vision sensing can also be found in the recent review \citep{2018DLW_review}.
	
	\section{Methodologies and Materials} 
	\subsection{Network Architecture}
	\subsubsection{Detection and Instance Segmentation}
	\begin{figure}
		\centering
		\includegraphics[width=.8\textwidth]{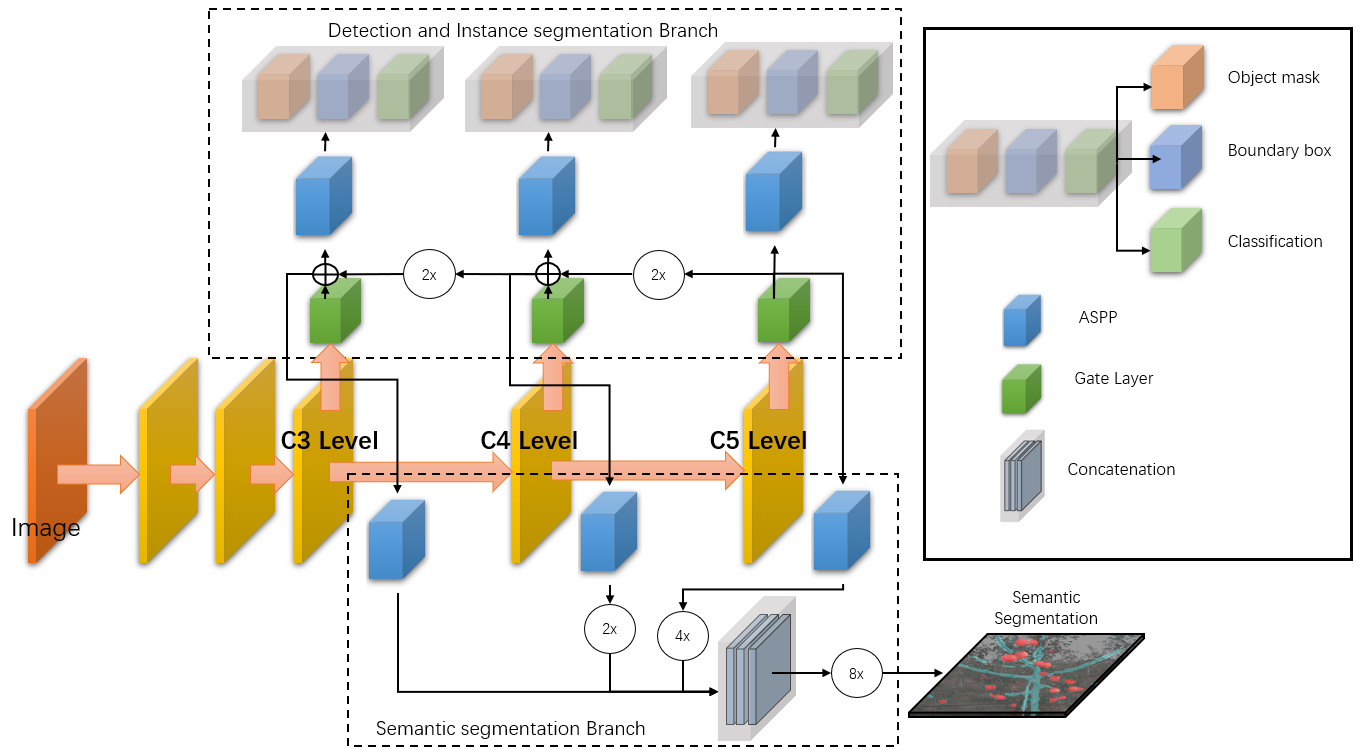}
		\caption{Network architecture of the DaSNet-V2.DaSNet-V2 is comprised by detection and instance segmentation branch and semantic segmentation branch. The detection and instance segmentation branch detect and segment the fruits, while the instance segmentation branch segments the branches.}
		\label{fig:network}
	\end{figure}
	\unskip
	DaSNet-V2 utilises one-stage detector to perform the detection and instance segmentation of fruits. In the previous work, two-stage detectors such as Mask-RCNN utilises the RPN and ROI pooling /align layer to predict and segment the corresponding area of objects on feature maps to perform the instance segmentation and detection. Recently, one-stage detector SPRNet applies ASPP to cover the corresponding area of objects and encode the multi-scale information to perform the instance segmentation on each grid of feature maps. DaSNet-V2 adopts the principle of the SPRNet, applying a mask branch on the output branch of each level of the gated-Feature Pyramid Network (FPN), to perform the instance segmentation and detection of fruits.
	
	\begin{figure}
		\centering
		\includegraphics[width=\textwidth]{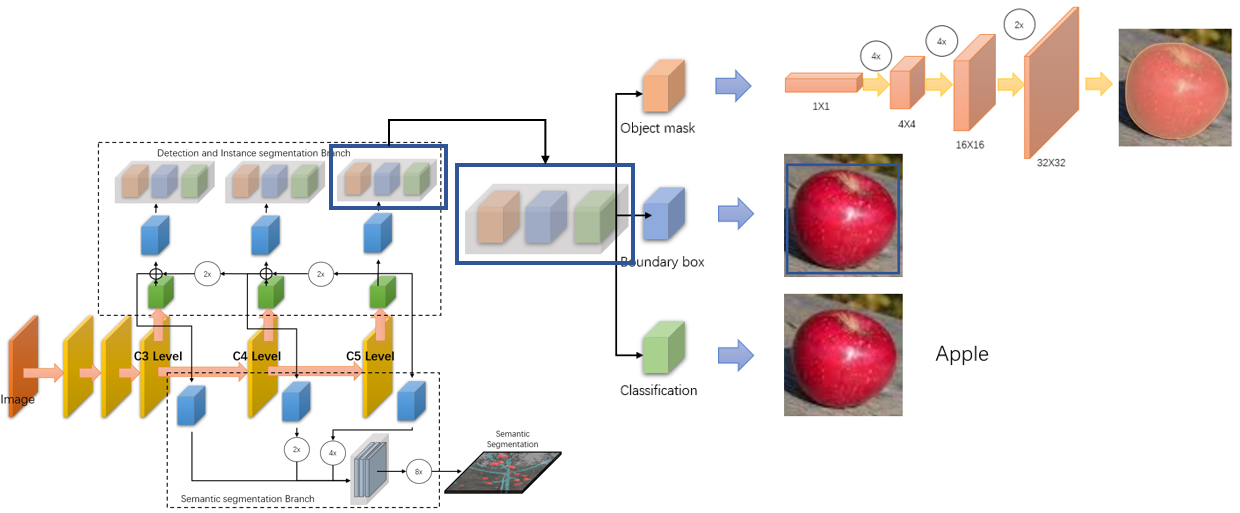}
		\caption{Design of the output branch of the DaSNet-V2. The output branch of the DaSNet-V2 comprises three subnets for object mask segmentation, object boundary box regression, and objects classification.}
		\label{fig:out_branch}
	\end{figure}
	
\begin{figure}
		\centering
		\includegraphics[width=\textwidth]{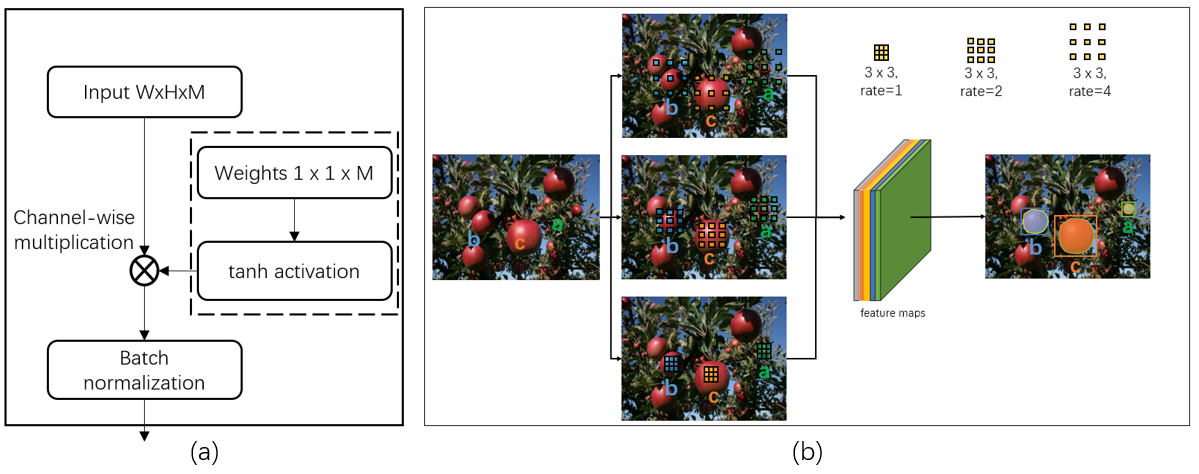}
		\caption{(a) Design of the gate layer in the gated-FPN of the DaSNet-V2. (b) Design of the ASPP which is utilised in the DaSNet-V2.}
		\label{fig:utils}
	\end{figure}

	The architecture of the DaSNet-V2 is shown in Figure \ref{fig:network}. DaSNet-V2 utilises a 3-level gated-FPN structure to fuse the information from the different level of feature maps. The architecture of the gated-FPN is shown in Figure \ref{fig:utils}. For feature maps from the different level, gated-FPN utilise a channel-wise multiplication layer with tanh activation to choose and re-adjust feature maps which are passed into the gated-FPN. Gated-FPN applied in the DaSNet-V2 receives feature maps from the C3, C4 and C5 level of the backbone and the fusion of feature maps between different level of gated-FPN is performed by using adding operation.
	
	On each level of the gated-FPN in the DaSNet-V2, an Atrous Spatial Pyramid Pooling (ASPP) is utilised to encode the multi-scale information of objects on feature maps into the output tensor. The applied ASPP utilise three dilation convolution kernels with dilation rate equal to 1, 2 and 4 and a 1 $\times$ 1 convolution kernel to cover the corresponding area of objects within feature maps. The output tensor of each ASPP is used to predict the class, boundary box, and mask of the objects, which is shown in Figure \ref{fig:out_branch}. The mask of objects is reconstructed by up-sampling of the feature tensor on each grid of feature maps, from the size of 1 $\times$ 1 $\times$ N (N is the channel number of the corresponding feature map) to the size of 32 $\times$ 32 $\times$ 2. The reconstructed object masks will be rescaled to the boundary box size based on the prediction of the DaSNet-V2. Each level of gated-FPN in the DaSNet-V2 has two preset anchor boxes, as experiment results show that six preset anchor boxes can properly predict and cover the range of possible shape of fruits in orchards.
	
	\subsubsection{Semantic Segmentation}
	The detection branch of the DaSNet-V2 detects and segments the fruits within images to locate and track the targets. However, such information is limited to guide the manipulator to perform a successful detachment of the target fruit, as there are many other factors presented within the working environment, such as densely arranged branches. Such factors lead to the failure of the detachment operation and even damage the gripper and manipulator system. To provide more information about the working scene, a semantic segmentation branch is developed and utilised in the DaSNet-V2 to perform the segmentation of branches.
	
	In the DaSNet-V1, three different architecture designs of semantic segmentation branch were developed and evaluated. The architecture of the semantic segmentation branch of the DaSNet-V2 follows the design strategy of the DaSNet-V1, which is shown in Figure \ref{fig:network}. Semantic segmentation branch receives feature maps from the C3, C4, and C5 level of the gated-FPN of the DaSNet-V2. To keep the consistency of size of feature tensors, feature maps from C3, C4 are 4 $\times$ and 2 $\times$ upsampled to match the size of the feature map from C5, respectively. The fusion between feature maps is achieved by using the concatenate operation, and three convolution layers are further applied after concatenating operation to process the feature information within feature maps. The output tensor of the semantic segmentation branch is 8 $\times$ upsampled to match the size of the input image. Different from the DaSNet-V1, the semantic segmentation branch of the DaSNet-V2 is designed to segment branches and no longer applied to segment the fruit, since the segmentation of fruits has been included in the detection and instance segmentation branch. 
	
	\subsection{3D Visualisation}
	\begin{figure}
		\centering
		\includegraphics[width=\textwidth]{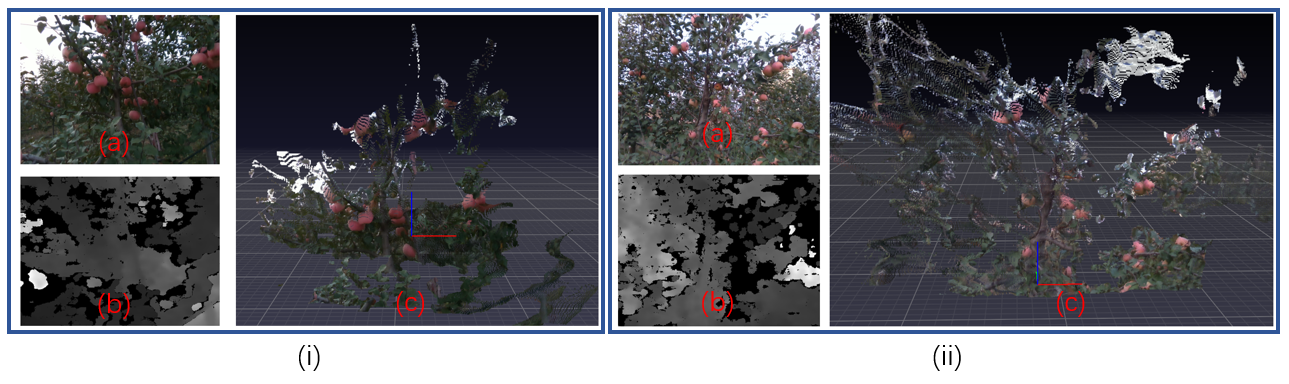}
		\caption{(a),(b), and (c) of figure (i) and (ii) are the RGB images, depth images, and cloud point in 3D space, respectively.}
		\label{fig:3d}
	\end{figure}
	In the traditional fruit orchards, the arrangement of the trees, branches and fruits are presented in random and complex behaviours. Such arrangement of branches and fruits could affect the performance of the harvesting robot to a large extent. Densely arranged branches may obstruct the path of the detachment and damage the manipulator and gripper system. Meanwhile, densely arranged fruits and different types of the stem-branch joint of fruits may also affect the success rate of fruit detachment. Some fruits can be easily picked while other fruits are difficult to be detached by the gripper. To provide more information for guiding the operation of the manipulator and gripper system, modelling and understanding the working scene from the 3D aspect is important.
	
	DaSNet-V2 can detect and segment each fruit and segment branches within the working scene. For the fruit class, different colours are assigned to the detected fruits to stand their shape and corresponding area. For the branch class, a unified colour is assigned to the pixels which have been classified as the branch. Other classes such as ground, fence and leaves are presented in black pixel. The segmentation of the leaves is not included in the DaSNet-V2 since previous in-field experiment results suggest that leaves only slightly affect the harvesting performance. In this work, PPTK tool-kit \citep{2018pptk} is utilised to visualise the 3D point cloud of the working scene, a sample of 3D visualisation of the orchard environment is shown in Figure \ref{fig:3d}.
	
	\subsection{Implementation details}
	\subsubsection{Training details}
	Data augmentation plays an important role in improving the performance of the trained model in deep learning. Training of the DaSNet-V2 follows the same strategy in training of the DaSNet-V1, a two-level object scale amplifier algorithm for minimising the uneven distribution of the object scale is utilised. Meanwhile, to cover more variance of objects appearance in fruit and branch class during the training, the ground truth of the instance segmentation and semantic segmentation are utilised to adjust the HSV, brightness and contrast of the pixels within the area of objects. Other augmentation measurements such as flip and rotation are also included in the training.
	
	Focal loss is utilised in the training of the detection of the DaSNet-V2 to balance the uneven distribution of the foreground class objects and background class objects, the cross-entropy is utilised in the training of the instance segmentation and semantic segmentation task. Adam-optimizer is applied in the network training, the learning rate and decay rate are set as 0.01 and 0.9 based on the previous experiment results, respectively.
	
	\subsubsection{Implementation details}
	The programming of the DaSNet-V2 is achieved by using slim tool-kit within the Tensorflow API-1.11 in Ubuntu 16.04. 3D visualisation of the point cloud is achieved by using PPTK tool-kit. The DaSNet-V2 is trained on the Nvidia GTX-1080Ti and be evaluated in the Intel CPU-i5 and Jetson TX2 in Ubuntu 16.04 and Nvidia GTX-1080Ti in Windows 10. Depth camera Intel RealSense D-435 is utilised in the experiment, and it is controlled by using ROS-kinetic in Ubuntu 16.04.
	
	\begin{figure}
		\centering
		\includegraphics[width=\textwidth]{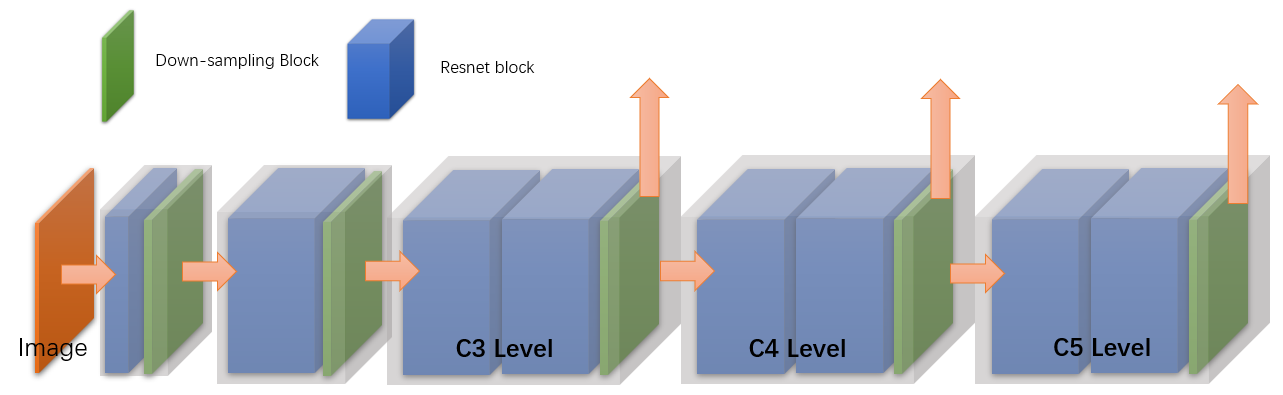}
		\caption{Architecture of the Lw-net. LW-net is comprised by 8 Resnet blocks and 5 down-sampling blocks. The design of the Resnet block and down-sampling block can be referred to the previous work \citep{2019DasNet-v1}.}
		\label{fig:LW-net}
	\end{figure}
	To ensure the computation availability and real-time performance of the DaSNet-V2 on the embedded computation device such as Jetson-TX2, a customised light-weight residual network 'LW-net' which was developed and evaluated in the DaSNet-V1 is utilised to serve as the backbone of the network. Meanwhile, other classification networks, including Resnet-101 and Darknet-53 are utilised to serve as the backbone of the network. The implemented code and ImageNet pre-trained weights of the Resnet-101 and Darknet-53 in Tensorflow were from the Github publicly code library, while the LW-net is pre-trained with Cifar-10/100 dataset. More details of the architecture and training of the LW-net is shown in Figure \ref{fig:LW-net}.
	
	\section{Experiment and Discussion}
	\subsection{Data collection and Evaluation Methods}
	The RGBD image data were collected from the apple orchard, which is located in Qingdao, China. The collection time of the image data was from 10:00 to 16:00 through the day by using the Intel RealSense D-435 depth camera. The image data was collected at the distance of 0.5-1.5m from the apple trees, which is the distance from the camera to the fruits in the harvesting. There are 300 RGB-D images and another 800 RGB images collected from the orchard. 500 out of these 1100 images were used to perform the training and rest of the images were used to perform the evaluation.  
	
	The evaluation of the DaSNet-V2 comprises three tasks, including the evaluation of the detection performance, instance segmentation quality and semantic segmentation quality. To evaluate the detection performance of the DaSNet-V2, $F_{1}$ score and Intersection of Union (IoU) are applied. $F_{1}$ combines the performance evaluation of the $recall$ and $precision$ of the detection; hence it has been widely applied as the evaluation index in many previous studies of the fruit detection. The expression of the $precision$, $recall$ and $F_{1}$ are presented as follow:
	\begin{equation}
	Precision=\frac{TruePositive(TP)}{TruePositive(TP)+FalsePositive(FP)}
	\end{equation}
	\begin{equation}
	Recall=\frac{TP}{TP+FalseNegative(FN)}
	\end{equation}
	\begin{equation}
	F_{1}=\frac{2\times\ Precision \times\ Recall}{Precision+Recall}
	\end{equation}
	The IoU measures the intersection area of the predicted object boundary box and the ground truth, to evaluate the location accuracy of the predicted boundary box of the prediction. On another hand, Mean Intersection of Union (MIoU) is used to evaluate the performance of the instance segmentation and semantic segmentation of the DaSNet-V2. The definition of the IoU and MIoU can be referred to the work \citep{2017semantic_segmentation_index}. 
	
	\subsection{Comparison to State of the Art}
	\subsubsection{Evaluation on Detection and Instance Segmentation}
	This experiment compares the detection performance between the DaSNet-V2 and the DaSNet-V1, YOLO-V3, YOLO-V3(tiny), Faster-RCNN  and the Mask-RCNN. Meanwhile, the comparison of the performance in instance segmentation between the DaSNet-V2 and the Mask-RCNN is also included in the experiment. The evaluation results of the comparison between different models are shown in Table as follow.
	
	\begin{table}
		\centering
		\caption{Comparison of performance of detection and instance segmentation between different network models}
		\begin{tabular}{ccccccc}
			\hline
			\textbf{Index}&\textbf{Model}&$F_{1}$&IoU&MIoU&Time\\
			\hline
			1&LedNet (Resnet-101) &${0.834}$&${0.872}$&${-}$&${46ms}$\\
			\hline
			2&DaSNet-V1 (Resnet-101) &${0.834}$&${0.872}$&${-}$&${72ms}$\\
			\hline
			3&DaSNet-V2 (Resnet-101) &${0.856}$&${0.881}$&${0.862}$&${65ms}$\\
			\hline
			4&YOLO-V3 (Darknet-53) &${0.797}$&${0.843}$&${-}$&${45ms}$\\
			\hline
			5&YOLO-Tiny (Darknet-18) &${0.787}$&${0.836}$&${-}$&${30ms}$\\
			\hline
			6&Faster-RCNN (VGG-16) &${0.814}$&${0.863}$&${-}$&${145ms}$\\
			\hline
			7&Mask-RCNN (Resnet-101) &${0.835}$&${0.865}$&${0.871}$&${137ms}$\\
			\hline
		\end{tabular}
		\label{table:experiment_1}
	\end{table}

    \begin{table}
    	\centering
    	\caption{Comparison of performance of the DaSNet-V2 in detection and instance segmentation with different backbones}
    	\begin{tabular}{ccccccc}
    		\hline
    		\textbf{Index}&\textbf{Model}&$F_{1}$&IoU&MIoU&Time\\
    		\hline
    		1&DaSNet-V2 (LW-net) &${0.844}$&${0.868}$&${0.858}$&${30ms}$\\
    		\hline
    		2&DaSNet-V2 (Darknet-53) &${0.848}$&${0.872}$&${0.863}$&${45ms}$\\
    		\hline
    		3&DaSNet-V2 (Resnet-101) &${0.856}$&${0.881}$&${0.862}$&${65ms}$\\
    		\hline
    	\end{tabular}
    	\label{table:experiment_2}
    \end{table} 
	
	YOLO-V3 and Faster-RCNN are the representative works of the one-stage detector and the two-stage detector, respectively. YOLO-Tiny is the light-weight version of the YOLO-V3 network. LedNet and DaSNet-V1 are the previous work of the DaSNet-V2. DaSNet-V1 and LedNet adopt same architecture design on fruit detection while DaSNet-V1 further applied a semantic segmentation branch in the model to perform the semantic segmentation of fruits and branches. In the experiment results 1,2 and 3, the detection performance of the DaSNet-V2 is improved compared to the DaSNet-V1 and LedNet. Experiment results 3-7 compare the detection performance of the DaSNet-V2, YOLO and Faster/Mask-RCNN. DaSNet-V2 achieves similar performance on fruit detection compared to the Faster/Mask-RCNN when same backbone Restnet-101 is applied, and better performance on fruit detection compared to the one-stage detector YOLO-V3 and YOLO-V3(Tiny). On the performance of the instance segmentation of fruits, DaSNet-V2 achieves a similar score compared to the Mask-RCNN, which are 0.862 and 0.871, respectively. 
	
	From the experiment results, DaSNet-V2 shows a better detection performance compared to the One-stage detector YOLO-V3 and YOLO-V3(Tiny). Compared to the two-stage detector Mask-RCNN, DaSNet-V2 achieves a better performance on the detection and an equal performance on instance segmentation. The experiment result shows that ASPP can improve the detection performance of the model and ensure the working of the mask branch. In terms of the computational efficiency, one-stage detector YOLO and DaSNet-V2 are faster than the two-stage detector. The computational time of model to process an image by using YOLO-V3 (Detection), LedNet (Detection), DaSNet-V2 (Detection+Instance Segmentation), and Mask-RCNN (Detection+Instance Segmentation) are 45ms, 46ms, 65ms, and 137ms, respectively.

	\subsubsection{Evaluation on Semantic Segmentation}
	This experiment compares the performance of semantic segmentation between the DaSNet-V2, DaSNet-V1 and the FCN-8s. The evaluation results of the comparison between different models are shown in Table as follow.
	
	\begin{table}
		\centering
		\caption{Comparison of performance of the DaSNet-V2 in semantic segmentation with different backbones}
		\begin{tabular}{ccccccc}
			\hline
			\textbf{Index}&\textbf{Model}&MIoU$_{branch}$&Time\\
			\hline
			1&DaSNet-V1 (Resnet-101) &${0.772}$&${72ms}$\\
			\hline
			2&DaSNet-V2 (Resnet-101) &${0.802}$&${65ms}$\\
			\hline
			3&FCN-8s (Resnet-101) &${733}$&${61ms}$\\
			\hline
		\end{tabular}
		\label{table:experiment_3}
	\end{table}
  
    \begin{table}
    	\centering
    	\caption{Comparison of performance of semantic segmentation between different network models}
    	\begin{tabular}{ccccccc}
    		\hline
    		\textbf{Index}&\textbf{Model}&MIoU$_{branch}$&Time\\
    		\hline
    		1&DaSNet-V2 (LW-net) &${0.795}$&${30ms}$\\
    		\hline
    		2&DaSNet-V2 (Darknet-53) &${0.797}$&${45ms}$\\
    		\hline
    		3&DaSNet-V2 (Resnet-101) &${0.802}$&${60ms}$\\
    		\hline
    	\end{tabular}
    	\label{table:experiment_4}
    \end{table}

    From the experimental results shown in Table \ref{table:experiment_3}, the semantic segmentation performance of the DaSNet-V2 is improved compared to the DaSNet-V1, which are 0.802 and 0.772, respectively. Compared to the DaSNet-V1, DaSNet-V2 utilised new architecture design of the ASPP and new augmentation methods, which can improve the generalisation of the network. Compared to the FCN-8s, the score of semantic segmentation achieved by DaSNet-V2 is 7$\%$ higher, which are 0.733 (FCN-8s) and 0.802 (DaSNet-V2), respectively. In terms of the computational efficiency, the computational time of DaSNet-V2 to process an image is slightly less than DaSNet-V1, as the architecture design of the gated-FPN in DaSNet-V2 is simplified and optimised compared to the DaSNet-V1.
    
    Table \ref{table:experiment_4} compares the performance of the DaSNet-V2 in semantic segmentation of branch with different backbones. From the experiment results, DaSNet-V2 with Resnet-101 outperforms in the comparison. The DaSNet-V2 with Darknet-53 and LW-net achieves similar performance, which is 0.797 and 0.795, respectively. In terms of the computational efficiency, DaSNet-V2 with LW-net requires 30ms for an image, while the DaSNet-V2 with Darknet-53 and Resnet-101 requires 45ms and 60ms to process an image, respectively. 
    
	\subsection{Visual Sensing in Orchards}
	\begin{figure}
		\centering
		\includegraphics[width=\textwidth]{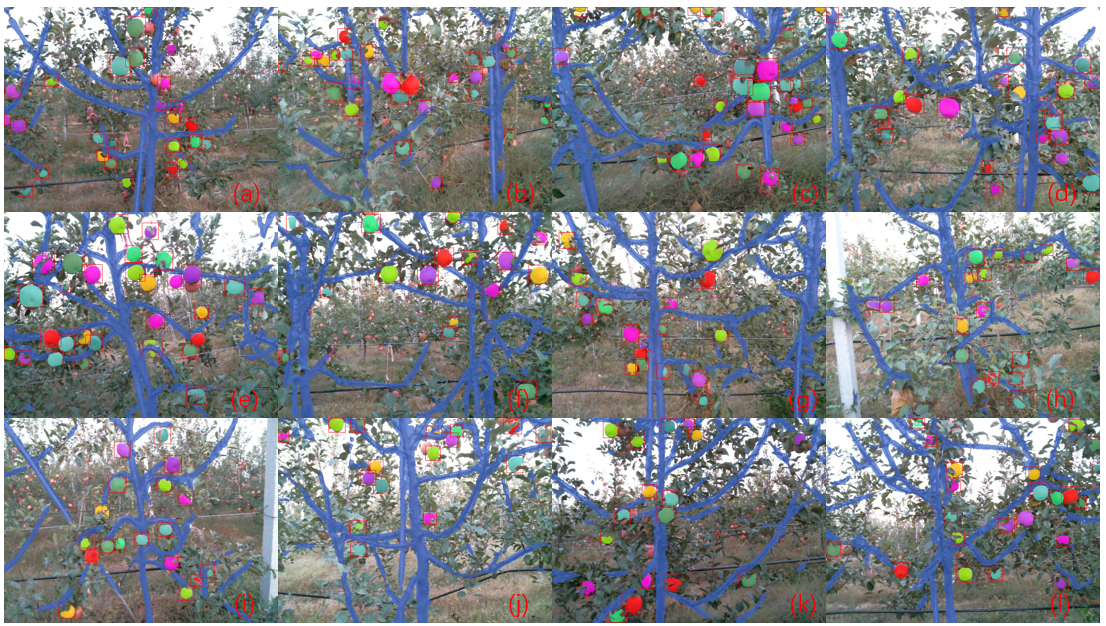}
		\caption{Detection and segmentation results in the apple orchard by using the DaSNet-V2.}
		\label{fig:seg}
	\end{figure}
	In traditional orchards, there are several issues that may affect the performance of automatic harvesting. Firstly, overlapping of fruits or occlusion between fruits, leaves and branches can affect the accuracy of the detection. Secondly, 'free-growth' and densely arranged branches can obstruct the detach path of the manipulator. This experiment evaluates the performance of the DaSNet-V2 in the real environment within the apple orchards. The experiment results, which are processed by using the DaSNet-V2 are shown in Figure \ref{fig:seg}.
	
	\begin{table}
		\centering
		\caption{Comparison of performance of semantic segmentation between different network models}
		\begin{tabular}{ccccccc}
			\hline
			\textbf{Device}&\textbf{Model}&\textbf{Weight Size}&\textbf{Time}\\
			\hline
			GTX-1080Ti &DaSNet-V2 (LW-net) &${8.1}$ M&${30ms}$\\
			\hline
			GTX-1080Ti &DaSNet-V1 (LW-net) &${12.8}$ M&${32ms}$\\
			\hline
			GTX-1080Ti &YOLO-Tiny (Darknet-18) &${35.4}$ M&${30ms}$\\
			\hline
			GTX-1080Ti &Faster-RCNN (VGG-16) &${533}$ M&${145ms}$\\
			\hline
			GTX-1080Ti &Mask-RCNN (Resnet-101) &${244}$ M&${137ms}$\\
			\hline
			Jetson-TX2 &DaSNet-V2 (LW-net) &${8.1}$ M&${278ms}$\\
			\hline
			Jetson-TX2 &DaSNet-V1 (LW-net) &${12.8}$ M&${326ms}$\\
			\hline
			Jetson-TX2 &YOLO-Tiny (Darknet-18) &${35.4}$ M&${265ms}$\\
			\hline
			Jetson-TX2 &Faster-RCNN (VGG-16) &${533}$ M&${1.3 s}$\\
			\hline
			Jetson-TX2 &Mask-RCNN (Resnet-101) &${244}$ M&${1.21s}$\\
			\hline
		\end{tabular}
		\label{table:experiment_5}
	\end{table}

	From the figures shown in results, DaSNet-V2 can accurately detect the fruit and segment the corresponding area of fruits from the background. Meanwhile, the segmentation results of the branch are also presented in fine and smooth detail. Also, DaSNet-V2 shows a robust performance to segment the fruits in the overlapping and occlusion conditions. Considering limited computation resource is available in orchard environments, the DaSNet-V2 is tested on the Jetson-TX2 to evaluate its computational availability. The computational efficiency of the DaSNet-V2 on GTX-1080Ti and Jetson-TX2 are listed in Table \ref{table:experiment_5}. It can be seen that the one-stage detector such as YOLO and DaSNet have obvious advantages to be deployed on the embedded computational device compared to the two-stage detector in terms of the computational efficiency. 
	
	\subsection{3D Visualisation of Environments}
	\begin{figure}[h]
		\centering
		\includegraphics[width=\textwidth]{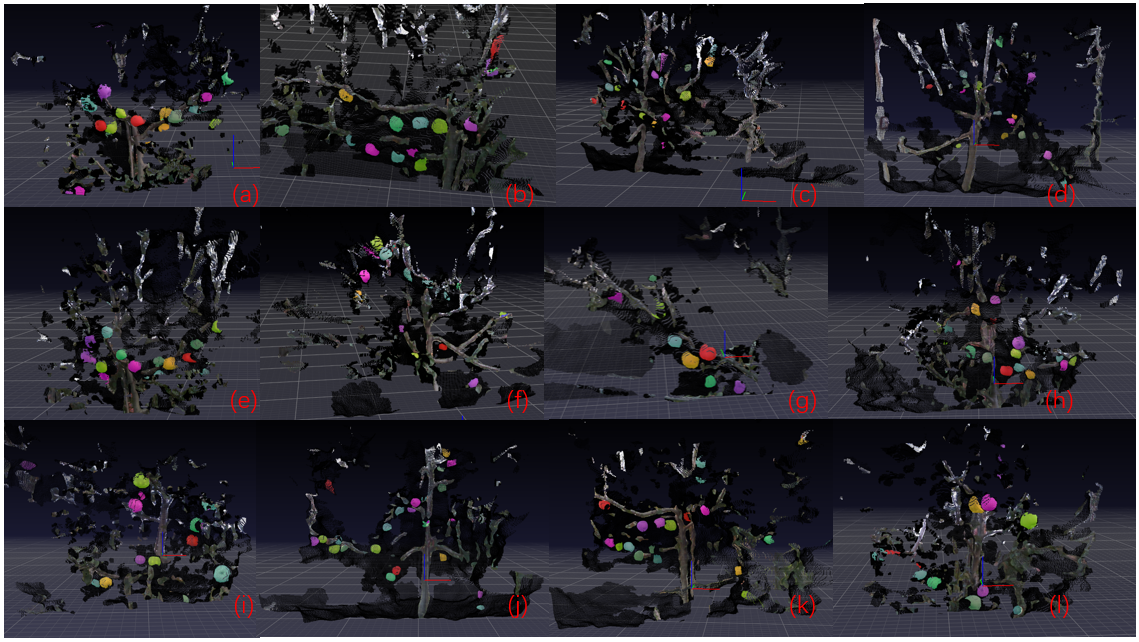}
		\caption{3D visualisation of the orchard environment by using the DaSNet-V2 and PPTK tool-kit. The apples are presented in colour mask, branches are presented by its original colour, and the other elements within the image are presented in black colour.}
		\label{fig:3dseg}
	\end{figure}
	Fruit detection and instance segmentation, and branch segmentation in 2D image space is a straightforward way of understanding the working environments in guiding the manipulator to detach the fruits. DaSNet-V2 is further evaluated to perform the work on the RGB-D images which are collected from the apple orchard. The 3D visualisation of the orchard environment, which are processed by using the DaSNet-V2 are shown in Figure \ref{fig:3dseg}.
	
	Orchard environment in 3D presentation form can clearly describe the working environment of the fruit harvesting robot. The orientation and shape of fruits, the structure of trees, and the possible orientation and location of the stem-branch joint of fruits can be seen or estimated from the 3D scene. Such information and scene can be further used to guide the detachment of the fruits or reconstruct and modelling the working environment of the fruit harvesting robot. From the 3D visualisation results which are shown in Figure, it can be seen that DaSNet-V2 can robustly and efficiently perform the multi-function work in the real orchard environment.
	
	\section{Conclusion and Future Work}
	In this study, a multi-function one-stage detector DaSNet-V2 was developed and evaluated. DaSNet-V2 comprises a detection and instance segmentation branch to perform the detection, segmentation, and localisation of the fruits and a semantic segmentation branch to perform the semantic segmentation of the branch within orchards. DaSNet-V2 adopts gated-FPN and ASPP to improve the detection performance of the model, a light-weight backbone LW-net was also developed and utilised in the DaSNet-V2 to improve the computational availability of the model on the embedded computational device. In the experiment, DaSNet-V2 shows accurate detection and segmentation performance. DaSNet-V2 with Resnet-101 achieved 0.856, 0.862, and 0.802 on detection, fruit segmentation, and branch segmentation, respectively. DaSNet-V2 with LW-net achieved 0.844, 0.858, and 0.795 on detection, fruit segmentation, and branch segmentation, respectively. The computational time of DaSNet-V2 with LW-net on GTX-1080Ti and Jetson-TX2 are 30ms and 265ms, respectively. From the experiment results, DaSNet-V2 showed robust and efficient performance to perform the vision sensing in orchards. Future work will focus on developing the orchard reconstruction algorithm based on the DaSNet-V2, corresponding control strategy for guiding the automatic robotic fruit harvesting will also be included.
	
	\section*{Acknowledgement}
	This work is supported by ARC ITRH IH150100006 and THOR TECH PTY LTD. We acknowledge Zijue Chen and Hongyu Zhou for their assistance in the data collection. And we also acknowledge Zhuo Chen for her assistance in preparation of this work. 
	
	\bibliographystyle{plainnat}
	\bibliography{Reference}
	\renewcommand\refname{References}   
	
\end{document}